\documentclass[runningheads]{llncs}
\usepackage{graphicx}
\usepackage{amsmath,amssymb}
\usepackage{color}

\usepackage{times}
\usepackage{epsfig}
\usepackage[]{algorithm2e}
\usepackage{subfigure}
\usepackage{multirow}
\usepackage{cite}
\usepackage{epstopdf}
\usepackage{pifont}
\usepackage[dvipsnames]{xcolor}
\usepackage{caption}
\usepackage{breqn}
\graphicspath{{figures/}}
\def\eqnvspace{{\vspace{-2mm}}}
\def\figvspace{{\vspace{-5mm}}}
\newcommand{\Paragraph}[1]{\vspace{-0mm} \noindent \textbf{#1} \hspace{0mm}}
\newcommand{\Section}[1]{\vspace{-2mm} \section{#1} \vspace{-2mm}}
\newcommand{\SubSection}[1]{\vspace{-2mm} \subsection{#1} \vspace{-2mm}}

\newcommand*{\boxedcolor}{red}
\makeatletter
\renewcommand{\boxed}[1]{\textcolor{\boxedcolor}{%
  \fbox{\normalcolor\m@th$\displaystyle#1$}}}
\makeatother

\makeatletter
  \newcommand\figcaption{\def\@captype{figure}\caption}
  \newcommand\tabcaption{\def\@captype{table}\caption}
\makeatother

\newenvironment{tight_itemize}{
\begin{itemize}
  \setlength{\topsep}{0pt}
  \setlength{\itemsep}{2pt}
  \setlength{\parskip}{0pt}
  \setlength{\parsep}{0pt}
}{\end{itemize}}

\begin{document}
\pagestyle{headings}
\mainmatter

\def\ACCV20SubNumber{558} 

\title{FAN: Feature Adaptation Network for Surveillance Face Recognition and Normalization} 
\titlerunning{Feature Adaptation Network for Surveillance FR and Normalization}
\author{Xi Yin\inst{1} \and Ying Tai\inst{2} \and Yuge Huang\inst{2} \and Xiaoming Liu\inst{1}}

\authorrunning{X. Yin et al.}

\institute{Michigan State University \email{yinxi.whu@gmail.com, liuxm@cse.msu.edu} \and
Youtu Lab, Tencent \email{\{yingtai,yugehuang\}@tencent.com}}

\maketitle

\begin{center}
\centering
\includegraphics[trim={0 0 0 0mm},clip,width=.99\textwidth]{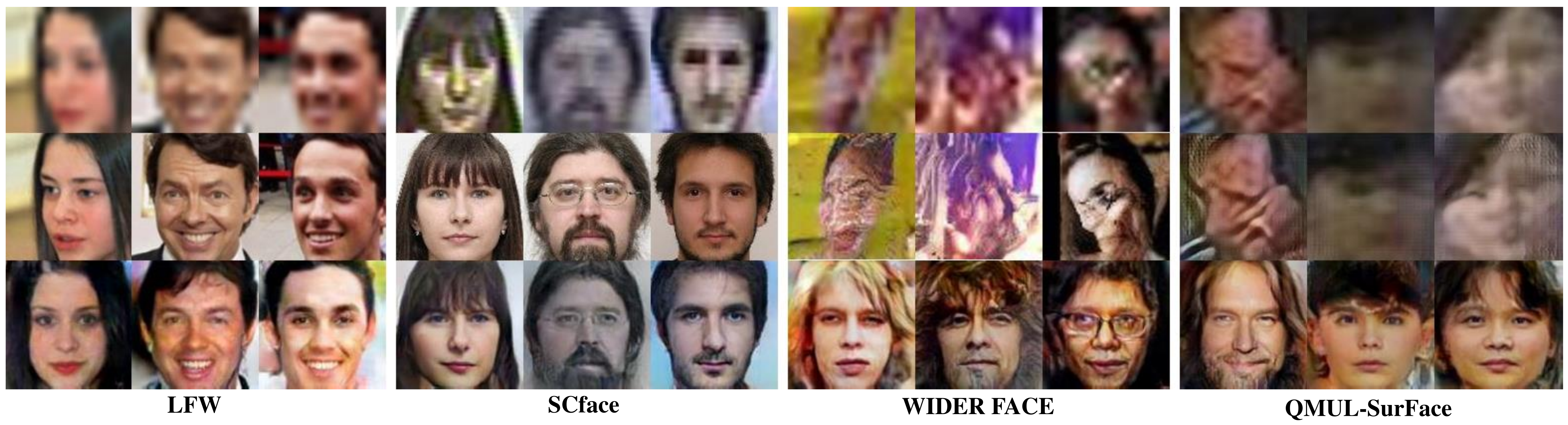} 
\vspace{-2mm}
\captionof{figure}{\small \textbf{Visual results on $4$ datasets.} 
Vertically we show input in row $1$ and our results in row $3$. For LFW and SCface, we show the ground truth and gallery images in row $2$. For WIDER FACE and QMUL-SurFace which do not have ground truth high-resolution images, we compare with two state-of-the-art SR methods: Bulat et al.~\cite{bulatyang2018learn} and FSRGAN~\cite{CT-FSRNet-2018} in row $2$.}
\label{fig:comp_fig1} 
\end{center}

\begin{abstract}
 This paper studies face recognition (FR) and normalization in surveillance imagery. Surveillance FR is a challenging problem that has great values in law enforcement. Despite recent progress in conventional FR, less effort has been devoted to surveillance FR. 
To bridge this gap, we propose a Feature Adaptation Network (FAN) to jointly perform surveillance FR and normalization. Our face normalization mainly acts on the aspect of image resolution, closely related to face super-resolution. However, previous face super-resolution methods require paired training data with pixel-to-pixel correspondence, which is typically unavailable between real-world low-resolution and high-resolution faces. FAN can leverage both paired and unpaired data as we disentangle the features into identity and non-identity components and adapt the distribution of the identity features, which breaks the limit of current face super-resolution methods. We further propose a random scale augmentation scheme to learn resolution robust identity features, with advantages over previous fixed scale augmentation. Extensive experiments on LFW, WIDER FACE, QUML-SurvFace and SCface datasets have shown the effectiveness of our method on surveillance FR and normalization. 
\end{abstract}

\Section{Introduction}
\label{section:1}
Surveillance Face Recognition (FR) is a challenge and important problem, yet less studied. 
The performance on conventional benchmarks such as LFW ~\cite{LFWTech} and IJB-A~\cite{klare2015pushing} have been greatly improved by state-of-the-art (SOTA) FR methods~\cite{wang2018cosface, wen2016discriminative, deng2019arcface}, which still suffer when applied to surveillance FR.
One intuitive approach is to perform Face Super-Resolution (FSR) on surveillance faces to enhance facial details.
However, existing FSR methods are problematic to handle surveillance faces, because they usually ignore the {\it identity} information and require {\it paired} training data.
In fact, preserving identity information is more crucial for surveillance faces than recovering other information, e.g., background, Pose, Illumination, Expression (PIE). 

In this work, we study surveillance FR and normalization. 
Specifically, given a surveillance face image, we aim to learn robust identity features for FR. Meanwhile, the features are used to generate a normalized face with enhanced facial details and neutral PIE. 
Our normalization is performed mainly on the aspect of resolution. 
While sharing the same goal as traditional SR, it differs in removing the pixel-to-pixel correspondence between the original and super-resolved images, as required by traditional SR. 
Therefore, we term it as face {\textit{normalization}}. 
For the same reason, we compare our work to previous FSR approaches, instead of prior normalization methods operating on pose~\cite{tran2017disentangled}, or expression~\cite{zhu2015high}.
To the best of our knowledge, this is the {\textit{first}} work to study surveillance face normalization.

We propose a novel Feature Adaptation Network (FAN) to jointly perform face recognition and normalization, which has three advantages over conventional FSR. 
$1$) Our joint learning scheme can benefit each other while most FSR methods do not consider the recognition task. 
$2$) Our framework enables training with both paired and unpaired data while conventional SR methods only support paired training. 
$3$) Our method simultaneously improves the resolution and alleviates the background and PIE from real surveillance faces while conventional methods only act on the resolution.
Examples in Fig.~\ref{fig:comp_fig1} demonstrate the superiority of FAN over SOTA FSR methods.

\begin{figure}
\begin{center}
\includegraphics[width=0.75\linewidth]{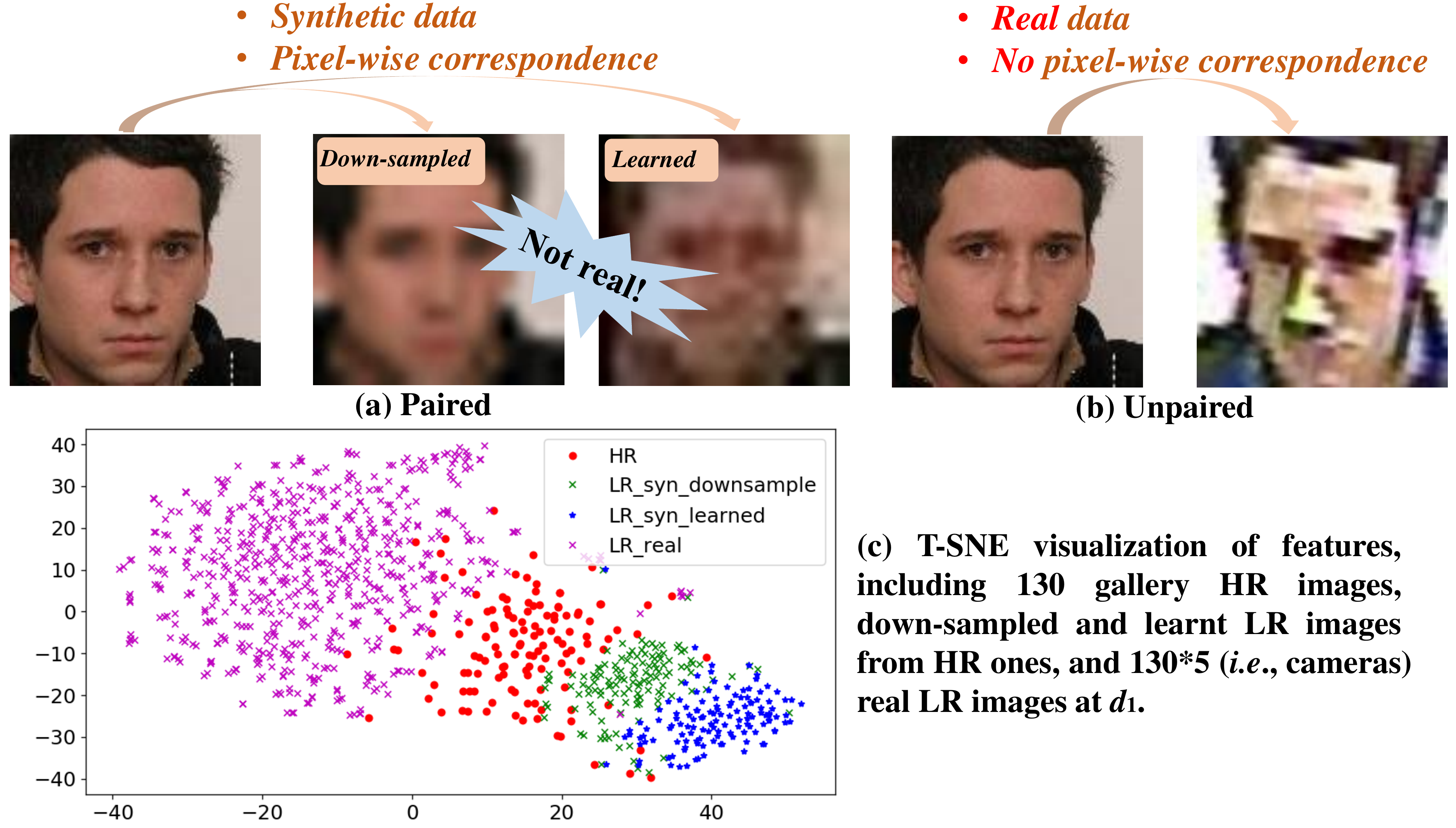}\vspace{-6mm}
\end{center}
\caption{\textbf{Paired vs. unpaired data from SCface~\cite{Grgic2011scface}.}
Synthetic paired data can be obtained by either down-sampling~\cite{CT-FSRNet-2018,CBN_ECCV16} or via a learned degradation mapping~\cite{bulatyang2018learn}.}
\vspace{-2mm}
\label{fig:paired_unpaired}
\end{figure}

Our FAN consists of two stages. 
In the first stage, we adopt disentangled feature learning to learn both identity and non-identity features mainly from high-resolution (HR) images, which are combined as the input to a decoder for pixel-wise face recovering. 
In the second stage, we propose feature adaptation to further facilitate the feature learning from the low-resolution (LR) images, by approximating the feature distribution between the LR and HR identity encoders. 
There are two advantages to use FAN for surveillance FR and normalization.
First, FAN focuses on learning disentangled identity features from LR images, which is better for FR than extracting features from super-resolved faces~\cite{tran2017disentangled,zhang2018facesr,wu2016j}.
Second, our adaptation is performed in the disentangled identity feature space, which enables training with unpaired data without pixel-to-pixel correspondence.
As shown in Fig.~\ref{fig:paired_unpaired}, the synthetic paired data used in prior works~\cite{CBN_ECCV16,CT-FSRNet-2018,bulatyang2018learn,wu2016j,zhang2018facesr,DRRN,MemNet_ICCV17, rad2019srobb} cannot accurately reflect the difference between real LR and HR in-the-wild faces, which is also observed in~\cite{cai2019toward}.

Furthermore, to better handle surveillance faces with unknown and diverse resolutions, we propose a {Random Scale Augmentation} (RSA) method that enables the network to learn all kinds of scales during training. 
Prior FSR~\cite{CT-FSRNet-2018,CBN_ECCV16,URDGN_ECCV16} methods either \textit{artificially} generate the LR images from the HR ones by simple {\textit{down-sampling}}, or \textit{learn} the degradation mapping via a Convolutional Neural Network (CNN). 
However, their common drawback is to learn reconstruction under \textit{fixed} scales, which may greatly limit their applications to surveillance faces.
In contrast, our RSA efficiently alleviates the constraint on scale variation.

In summary, the contributions of our work include:
\vspace{-2mm}
\begin{tight_itemize}
\item We propose a novel FAN to address surveillance face recognition and normalization, which is suitable for both paired and unpaired data.
\item We integrate disentangled feature learning to learn identity and non-identity features, which helps achieve face normalization for visualization, and \textit{identity preserving} for face recognition, simultaneously.
\item We propose a random scale augmentation strategy in FAN to learn various scales during training, which addresses the unknown resolutions of surveillance faces.
\item We achieve state-of-the-art performances on surveillance face datasets: WIDER FACE, QMUL-SurvFace and SCface, both quantitatively and qualitatively.
\end{tight_itemize}

\begin{table*}[t!]
\scriptsize
\caption{\small \textbf{Comparisons with previous state-of-the-art face super-resolution methods}.
Vis., Rec., Dis. and Frontal. indicate visualization, recognition, disentangled and frontalization, respectively. $A\rightarrow B$ refers to performing A first and then followed by B.}
\label{table:comparison}
\centering
\resizebox{\linewidth}{!}{
\begin{tabular}{c|ccccc|cc|}
\hline
\multirow{2}{*}{Method} & FSRNet~\cite{CT-FSRNet-2018} & Bulat et al.~\cite{bulatyang2018learn}   & S$^2$R$^2$~\cite{Henning2008facesr} & Wu et al.~\cite{wu2016j} & SICNN~\cite{zhang2018facesr} & \multirow{2}{*}{FAN (Ours)} \\
                        & (CVPR'$18$) & (ECCV'$18$) & (CVPR'$08$)   & (Arxiv'$16$) & (ECCV'$18$) &  \\
\hline
Deep Model      & $\surd$ & $\surd$ & $\times$ & $\surd$ & $\surd$ & $\surd$  \\ \hline
Applications	& Vis. & Vis. & Vis. $\&$ Rec. & Vis.$\&$ Rec. & Vis. $\&$ Rec. & Vis. $\&$ Rec. \\ \hline
Pipeline     & SR & SR & Features$\&$SR$\rightarrow$Rec. & SR$\rightarrow$Rec. & SR$\rightarrow$Rec. & Dis. Features$\rightarrow$SR$\&$Frontal.$\&$Rec. \\ \hline
Scale Factors	& $8$ & $4$ & $2$/$4$  & $4$ & $8$ & Random \\ \hline
Identity Preserving	& $\times$ & $\times$ & $\surd$ & $\surd$ & $\surd$  & $\surd$ \\ \hline
\multirow{2}{*}{Scenarios}         & CelebA/Helen & WIDER FACE & MPIE/FRGC  & LFW/YTF & LFW & SCface/QMUL-SurFace/ \\
                                   & (Easy)     & (Hard)     & (Easy)  & (Medium)  & (Medium)  & WIDER FACE (Hard) \\
\hline
\end{tabular}
}
\end{table*}

\Section{Related Work}
\label{section:2}
\Paragraph{Face Recognition}
Face recognition is a long-standing topic in computer vision. 
The performance has improved substantially due to the success of CNNs and large training sets~\cite{guo2016ms1m}. 
Most previous FR methods are focused on designing better loss functions to learn more discriminative features~\cite{wen2016discriminative, yin2019towards, deng2019arcface, wang2018cosface, Huang_2020_CVPR}. 
For example, Deng et al.~\cite{deng2019arcface} proposed ArcFace to introduce a margin in the angular space to make training more challenging and thus learn a more discriminative feature representation. 
Other methods are proposed to handle one or more specific variations in FR. 
For example, pose-invariant FR~\cite{tran2017disentangled, yin2018multi, zhao20183d} has been widely studied as pose is one major challenge in FR. 
With recent advance in FR, the performance on conventional benchmarks like LFW~\cite{LFWTech}, CFP~\cite{sengupta2016frontal}, and IJB-A\cite{klare2015pushing} are saturating. 

However, most previous FR methods fail to achieve satisfactory performance on surveillance FR~\cite{cheng2018qmulface}, which is a more challenging task that tackles unconstrained LR faces.
To address surveillance FR, one common approach is to learn a unified feature space for LR and HR images~\cite{Sheljar2011fsr, li2016fsr}.
Besides, face SR methods that preserve the identity information are another direction. 
E.g., Hennings-Yeomans et al.~\cite{Henning2008facesr} incorporated face features based prior in SR. 
Wu et al.~\cite{wu2016j} integrated a FR net after a standard SR net, and jointly learned a deep model for face hallucination and recognition.
More recently, Zhang et al.~\cite{zhang2018facesr} defined a super-identity loss to measure the identity difference within the hypersphere identity metric space.
Different from methods that performed recognition based on the recovered HR images~\cite{wu2016j,zhang2018facesr}, our method is more like~\cite{Henning2008facesr} that firstly learns the features, but differs in three aspects:
$1$) Our \textit{non-linear and compact} features learned from the deep CNNs are more powerful than the linear features in~\cite{wu2016j,zhang2018facesr}.
$2$) Our FAN focuses on disentangled identity features and thus can fully leverage both paired and unpaired data.
$3$) Our FAN well handles LR \textit{surveillance} faces.

\Paragraph{Face Normalization}
It is widely assumed in prior work~\cite{belhumeur1997eigenfaces, susskind2011modeling, chen2012bayesian} that the appearance of a face is influenced by two factors: identity and intra-class (or {\textit{non-identity}}) variation. 
Normally face normalization is a general task of generating an identity-preserved face while removing other non-identity variation including pose, expression, illumination, and resolution. 
Most prior works of face normalization have focused on specifically removing pose variation, i.e., face frontalization~\cite{tran2017disentangled, yin2017towards, huang2017beyond}, expression variation~\cite{amberg2008expression, zhu2015high}, or illumination variation~\cite{chen2006illumination}.
Other works~\cite{zhu2015high, qian2019unsupervised} perform pose and expression normalization.
In contrast, our work mainly focuses on resolution normalization by enhancing facial details, which also handles PIE implicitly. 
Motivated by the disentanglement-based face recognition approaches~\cite{tran2017disentangled, hadad2018two}, we incorporate disentangled feature learning for LR face normalization. 

Our work is close to face SR. 
Early face SR work~\cite{FH_Baker,StructuredFH_CVPR13,TBSFSR_CVPR15} adopt different types of machine learning algorithms.
For example, Baker and Kanade~\cite{FH_Baker} learned a prior on the spatial distribution of the image gradient for frontal faces.
Yang et al.~\cite{StructuredFH_CVPR13} assumed that facial landmarks can be accurately estimated from the LR face image, and incorporated the facial priors by using the mapping between specific facial components.
Recently, deep CNN has shown its superiority for face SR.
Zhou et al.~\cite{zhou2015facesr} proposed a bi-channel CNN for faces with large appearance variations.
Zhu et al.~\cite{SRCNN_PAMI16} super-resolved unaligned low-resolution faces in a task-alternating cascaded framework.
More recently, several works~\cite{CT-FSRNet-2018,URDGN_ECCV16} adopt Generative Adversarial Networks (GAN) to generate photo-realistic face images.
However, the above methods ignore the identity information during training, which is essential for human perception~\cite{zhang2018facesr} and downstream FR tasks.
In contrast, our method jointly learns surveillance face recognition and normalization, which is based on the disentangled identity information.
We compare with the most relevant FSR papers in Tab.~\ref{table:comparison}.

\begin{figure*}
\begin{center}
\includegraphics[trim=80 165 75 70, clip,width=0.95\textwidth]{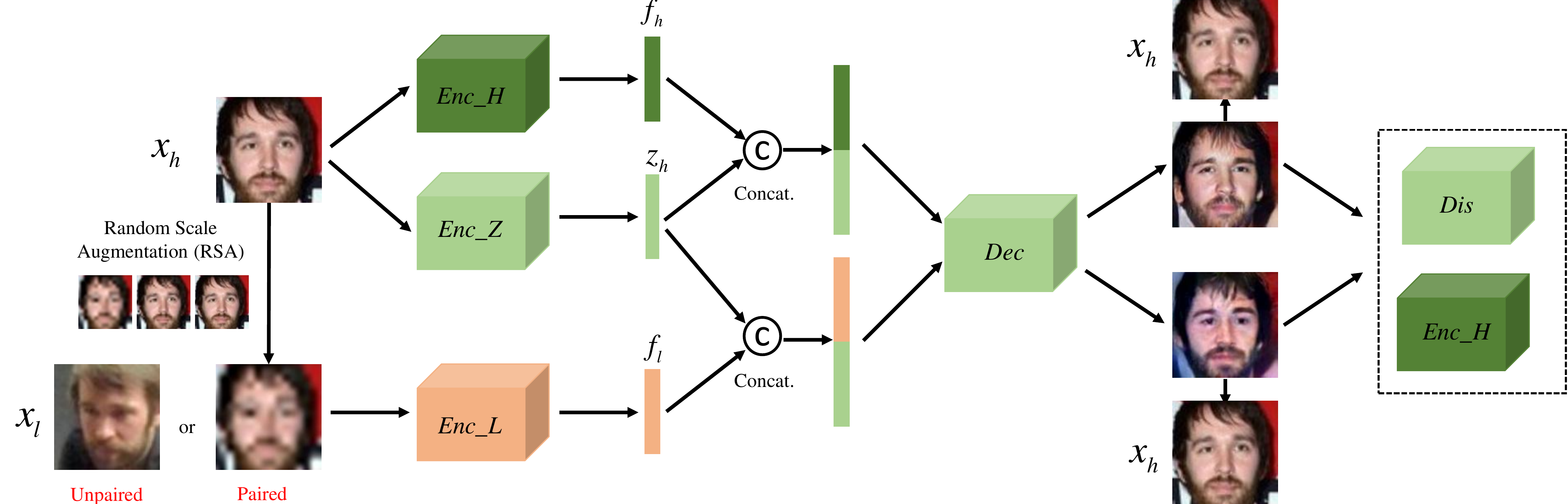}
\vspace{-2mm}
\figcaption{\textbf{Overview of FAN}. Green parts represent disentangled feature learning in stage one, where dark green and light green are the two steps. First, $Enc\_H$ (dark green) is a HR identity encoder that is pre-trained and fixed. Second, $Enc\_Z$, $Dec$, and $Dis$ are trained for feature disentanglement. Orange parts represent the feature adaptation in stage two where a LR identity encoder is learned with all other models (green) fixed.}
\label{fig:overview}
\figvspace
\end{center}
\end{figure*}

\Section{Feature Adaptation Network}
\label{section:3} 
In this section, we first give an overview of FAN (Section~\ref{section:3.1}) that consists of two stages. A feature disentanglement framework is introduced in Section~\ref{section:3.2}, which aims to disentangle identity features with other factors of variations. Then we propose the feature adaptation with random scale augmentation in Section~\ref{section:3.3} to learn resolution-robust identity features from both paired and unpaired training data. 

\SubSection{Framework Overview} \label{section:3.1}
The goals of our work are two-folds: 1) resolution-robust face recognition; 2) identity-preserved and resolution-enhanced face normalization. We propose to learn a disentangled representation to achieve both tasks. Performing face normalization from disentangled features enables identity supervision on both the disentangled features and the normalized faces, in contrast to previous SR work where identity supervision is only applied to the super-resolved faces. Such identity supervisions allow us to leverage real {\it unpaired} HR and LR faces without pixel correspondence. This is crucial for tackling surveillance scenario where paired images are usually unavailable in a large scale.

As shown in Fig.~\ref{fig:overview}, our method consists of two stages: disentangled feature learning (green components) and feature adaptation (orange components). 
Feature disentanglement has been successfully applied to face recognition and face synthesis~\cite{tran2017disentangled, liu2018exploring}. 
A disentangled representation is not only generative for face synthesis but also discriminative for face recognition.
In this stage, we train our feature disentanglement framework with HR faces. 
A face image is encoded into identity and non-identity features, which are combined to generate the input image.

In the second stage, we fix all models in the disentanglement framework and perform feature adaptation with HR-LR input images that can be either {\it paired} or {\it unpaired}. 
A LR feature encoder is learned to extract discriminative identity features from LR faces. 
The disentanglement framework provides strong supervisions in the feature adaptation process. 
To achieve resolution-robust recognition, we propose Random Scale Augmentation (RSA) to overcome the drawbacks of fixed-scale SR in previous work~\cite{bulatyang2018learn,CT-FSRNet-2018}.

\SubSection{Disentangled Feature Learning} \label{section:3.2}
Our feature disentanglement framework consists of five modules: an identity feature encoder $Enc\_H$, a non-identity feature encoder $Enc\_Z$, a decoder $Dec$, a linear classifier $FC$ (omitted from Fig.~\ref{fig:overview} for clarity), and a discriminator $Dis$.
To disentangle identity features from the non-identity variations, we perform a two-step feature disentanglement motivated by~\cite{hadad2018two}, but differs in three aspects. 

In the first step, a state-of-the-art face recognition model is trained with HR and down-sampled LR images using standard softmax loss and m-$L_2$ regularization~\cite{yin2019feature}.
We denote the trained feature encoder as $Enc\_H$, which remains fixed for all later stages to provide encoded identity features ${\bf{f}}_h = Enc\_H({\bf{x}}_h)$ from a HR input ${\bf{x}}_h$. In the second step, we aim to learn non-identity features ${\bf{z}}_h = Enc\_Z({{\bf{x}}_h})$ by performing adversarial training and image reconstruction. 

The first difference to~\cite{hadad2018two} is the loss for ${\bf{z}}_h$. ~\cite{hadad2018two} minimizes the identity classification loss, which we found to be unstable during training as it is unbounded. Instead, we propose to enforce the non-identity features to be evenly classified to all identities to make ${\bf{z}}_h$ identity {\textit{unrelated}}~\cite{liu2018exploring}.
\eqnvspace
\begin{equation}
    L_{z} = || FC({\bf{z}}_h) - {\bf{y}}_z ||_2^2,
\label{eqn:z}
\end{equation}
where ${\bf{y}}_z = [\frac{1}{N_D},\dots,\frac{1}{N_D}]\in\mathbb{R}^{N_D}$ and $N_D$ is the total number of identities in the training set. 
The gradient of this loss is used to  update only $Enc\_Z$ but not $FC$.

The disentangled features are combined to generate a face image ${\bf{x}}_{h}' = Dec({\bf{f}}_h, {\bf{z}}_h)$ with the goal of recovering the input: $L_{dec} = || {\bf{x}}_{h}' - {\bf{x}}_h ||_2^2$.
As ${\bf{f}}_h$ is discriminative for face recognition, the non-identity components will be discarded from ${\bf{f}}_h$ in the first step. 
The reconstruction will encourage $Enc\_Z$ to encode non-identity features ${\bf{z}}_h$ that is complimentary to ${\bf{f}}_h$ in order to recover the input face. 

The second difference to~\cite{hadad2018two} is that we employ an identity similarity regularization and GAN-based discriminator loss to impose identity similarity and improve visual quality of the generated faces. Specifically, for identity loss we use $Enc\_H$ to extract features and regularize the feature distance: $L_{id} = || Enc\_H({\bf{x}}_{h}') - {\bf{f}}_h ||_2^2$.
For GAN-based discriminator loss, we use standard binary cross entropy classification loss that is omitted here for clarity. 

\begin{figure}
\begin{center}
\includegraphics[trim=50 158 340 140, clip,width=0.6\textwidth]{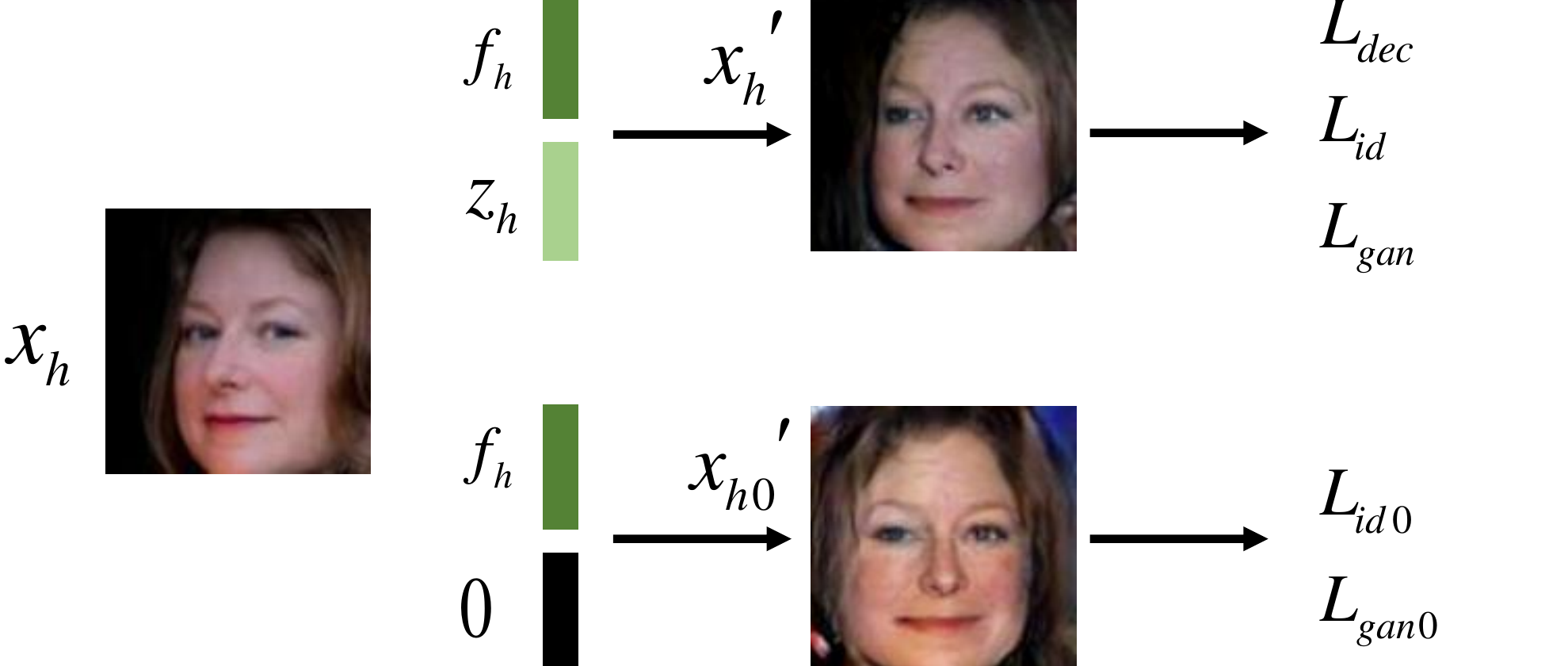}
\vspace{-2mm}
\figcaption{\textbf{Our feature disentanglement learning performs two kinds of reconstructions}. Top row denotes the reconstruction from the identity and non-identity features where we supervise on both pixel domain reconstruction ($L_{dec}$) and feature level regularization ($L_{id}$ and $L_{gan}$). Bottom row denotes the reconstruction from the identity features where only feature-level regularization are used.}
\label{fig:disentangle}
\figvspace
\end{center}
\end{figure}

The third difference to~\cite{hadad2018two} is that we perform an additional reconstruction to generate a face image from the identity features alone: ${\bf{x}}_{h0}'=Dec({\bf{f}}_h, {\bf{0}})$ where $\bf{0}$ represents a vector of $0$ that is of the same dimension as ${\bf{z}}_h$. 
As the non-identity part is given as $\bf{0}$, we expect the generated face to be an identity-preserved and {\it normalized} face without variations in non-identity factors such as PIE. As there is no ground-truth target image, we impose identity and GAN loss respectively. This process is illustrated in Fig.~\ref{fig:disentangle}.

The additional reconstruction has two benefits. 
First, it encourages the identity features alone to synthesize an identity-preserved face, which in turn prevents the non-identity features to encode identity information and results in better disentanglement.
Second, the ability of $Dec$ to reconstruct an identity-preserved face from the identity features alone is useful to enhancing facial details during inference (Section~\ref{section:3.3}).

\SubSection{Paired and Unpaired Feature Adaptation} \label{section:3.3}
In the first stage, we have learned disentangled representation and image reconstruction from HR images. The reason to train $Dec$ with HR only is to force $Dec$ to generate HR images, which is the goal to enhance resolution for face normalization. However, this framework will not work well for LR inputs. Therefore, we propose a feature adaptation scheme to learn a LR Encoder $Enc\_L$ for LR face recognition and normalization. We aim to learn a feature extractor that works well for input faces with various resolutions. 

\Paragraph{Training with paired data} In conventional FSR, it is common to down-sample the HR faces to LR versions with a few fixed scales and use the HR images to supervise FSR. 
Such methods cannot well handle the various resolutions in real-world surveillance imagery. 
To solve this issue, we propose Random Scale Augmentation (RSA).
Given a HR input ${\bf{x}}_h\in{\mathbb{R}}^{N_h\times N_h}$, we down-sample the image to a random resolution to obtain ${\bf{x}}_l\in{\mathbb{R}}^{K\times K}$, where $K\in [N_l, N_h]$ and $N_l$ is the lowest pixel resolution we care about ({\emph{e.g.}} $8$).
We call the HR images and down-sampled LR counterparts as paired data as they have pixel-to-pixel correspondence.  

\Paragraph{Training with unpaired data} Unpaired data represents HR and LR face images from the same subject but do not have pixel-wise correspondence. 
As shown in Fig.~\ref{fig:paired_unpaired}, the distribution of paired data is very far away from that of unpaired data. 
However, conventional FSR cannot take advantage of such unpaired data. 
Fortunately, FAN can well handle such case as we conduct face normalization from the disentangled features. 
As shown in Fig.~\ref{fig:overview}, FAN is suitable for both paired and unpaired training because it adapts the feature distributions between LR and HR images. 
We perform both feature-level and image-level similarity supervisions. 

Specifically, given a LR face image ${\bf{x}}_l$ that is obtained from either random down-sampling or unpaired source, we resize ${\bf{x}}_l$ to the same dimension as ${\bf{x}}_h$ using bicubic interpolation.
Then we use $Enc\_L$ to extract identity features ${\bf{f}}_l$, which is regularized to be similar to the disentangled features of the corresponding HR input image:
\eqnvspace
\begin{equation}
L_{enc} = || Enc\_L({\bf{x}}_l) - Enc\_H({\bf{x}}_h) ||_2^2.
\end{equation}

\vspace{-2mm}
This feature-level regularization adapts the features of LR images to the HR images in the disentangled feature space. 
The second regularization is defined in the recovered face image space. 
Recall that in the first stage $Dec$ is trained to generate a HR image from the identity and non-identity features of a HR input face.
If $Enc\_L$ can encode identity-preserved features, such features can replace the original HR identity features ${\bf{f}}_h$ to recover the HR input face. 
Thus, we impose an image-level regularization: 
\eqnvspace
\begin{equation}
    L_{enc\_dec} = || Dec({\bf{f}}_l, Enc\_Z({\bf{x}}_h)) - {\bf{x}}_h||_2^2.
\end{equation}

\vspace{-2mm}
As the non-identity features encoded from the HR image contributes to generating the output, the original HR can be used as the target for supervision.
This formulation is fundamentally different to all previous face SR methods that cannot impose pixel-wise supervision from unpaired data. 

Both feature-level and image-level regularization will enforce $Enc\_L$ to learn robust identity features ${\bf{f}}_l$. 
By varying the resolutions of the inputs, ${\bf{f}}_l$ is resolution robust. 
We also encourage the generated output to be realistic and identity preserving using the pre-trained discriminator $Dis$ and $Enc\_H$. 
The detailed training steps are summarized in Tab.~\ref{tab:train}. 
First, we train $Enc\_H$ using HR and down-sampled LR images in stage $1.1$. 
Second, we train feature disentanglement using HR images only in stage $1.2$ by fixing $Enc\_H$.
Third, we train $Enc\_L$ with all other models fixed in stage $2$. 

\Paragraph{Inference} 
We extract identity features ${\bf{f}}_l = Enc\_L({\bf{x}}_l)$ from a LR input for face recognition, 
and can further perform face normalization via $Dec({\bf{f}}_l, \bf{0})$ by setting the non-identity component to $\bf{0}$.
Thus we do not require HR images during inference.

\begin{table}[t]
\begin{center}
\footnotesize
\tabcaption{\textbf{Detailed training steps of FAN}. The first stage ($1.*$) involves a two-step disentanglement learning process. The second stage ($1.2$) is the feature adaptation process.}
\label{tab:train}
\begin{tabular}{@{}l@{\hspace{1mm}}|@{\hspace{1mm}}c@{\hspace{1mm}}|@{\hspace{1mm}}c@{\hspace{1mm}}|@{\hspace{1mm}}c@{}}
\hline
Stage & Input & Training Models & Fixed Models \\ \hline
$1.1$ & HR+LR    & $Enc\_H$   & $-$  \\ 
$1.2$ & HR    & $Enc\_Z$, $FC$, $Dec$, $Dis$  & $Enc\_H$\\ \hline
$2$   & HR+LR   & $Enc\_L$ & $Enc\_H$, $Enc\_Z$, $Dec$, $Dis$ \\ \hline 
\end{tabular}
\end{center}
\figvspace
\end{table}

\Section{Experiments}\label{section:4}
\SubSection{Implementation Details}\label{section:4.1}
\Paragraph{Datasets}
We conduct extensive experiments on several datasets including refined MS-Celeb-1M (MS1M)~\cite{deng2019arcface} for training, and LFW~\cite{LFWTech}, SCface~\cite{Grgic2011scface}, QMUL-SurvFace~\cite{cheng2018qmulface} and WIDER FACE~\cite{yang2016widerface} for testing.
Refined MS1M, a cleaned version of the original dataset~\cite{guo2016ms1m}, contains $3.8$M images of $85$K identities.
For LFW, we down-sample the $6,000$ face pairs to low resolution and adopt the standard evaluation protocol. 
SCface consists of HR (gallery) and LR (probe) face images of $130$ subjects. 
Following~\cite{lu2018sr}, $50$ subjects are used for fine-tuning and $80$ subjects are for testing. We conduct face identification where HR images are used as the gallery set and LR images with different resolutions (captured at three distances: $4.2$m for $d_1$, $2.6$m for $d_2$ and $1.0$m for $d_3$) form the probe set. QMUL-SurvFace consists of very-low resolution face images captured under surveillance cameras, and is used for face verification. 

\Paragraph{Training Setting}
Following~\cite{zhang2016mtcnn}, we use five facial landmarks (eye centers, nose tip and mouth corners) to align a face image to $128\times 128$.
We uniformly re-size the input LR images to a fixed size of $128\times 128$ by bicubic interpolation, which makes our method suitable for the proposed RSA strategy. 
Our framework is implemented with the Torch7 toolbox~\cite{Torch7}.
Our $Enc\_H$ and $Enc\_L$ are based on ResNet-$50$~\cite{ResNet_CVPR16}, $Enc\_Z$, $Dec$, and $Dis$ are similar to~\cite{tran2017disentangled}. 
We train stages $1.1$ and $1.2$ with a learning rate of $2e^{-4}$ for $12$ and $8$ epochs respectively. 
Stage $2$ is trained with a learning rate of $2e^{-5}$ for $6$ epochs. We use Adam optimization~\cite{kingma2014adam}. 
For SCface experiments, we finetune $Enc\_L$ with refined-MS1M (\textit{paired}) and SCface (\textit{unpaired}) training set for $1,000$ iterations with a learning rate of $1e^{-5}$. 

\begin{figure}
\begin{center}
\begin{tabular}{@{}c@{}c@{}c@{}c@{}c@{}c@{}c@{}}
(a) &
\includegraphics[trim=1152 384 0 0, clip, width=0.11\textwidth]{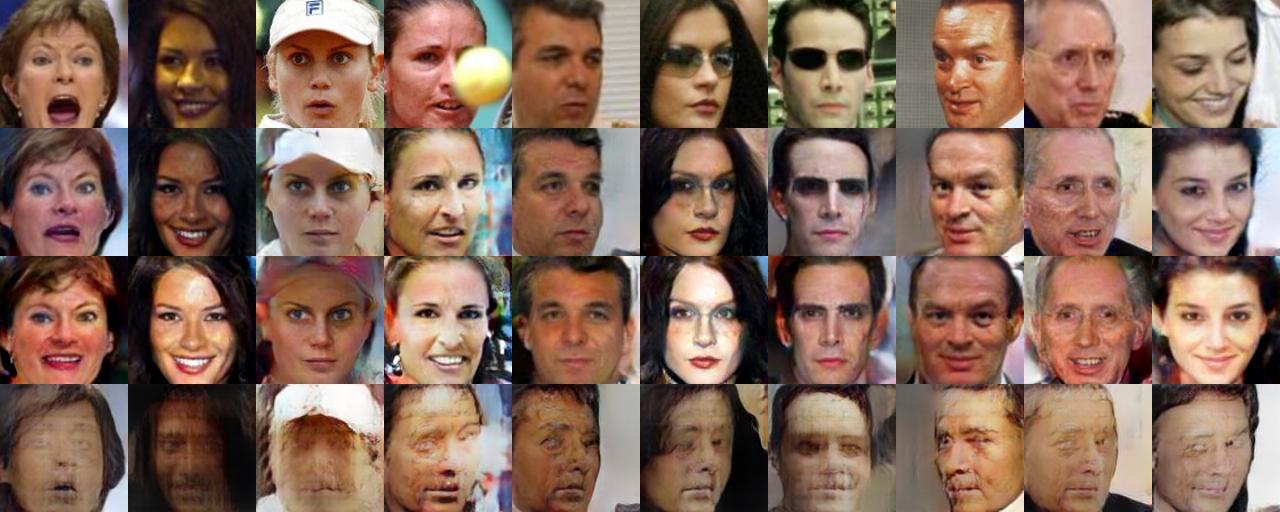} &
\includegraphics[trim=896 384 256 0, clip, width=0.11\textwidth]{figures/lfw_disentangle.jpg} &
\includegraphics[trim=128 384 1024 0, clip, width=0.11\textwidth]{figures/lfw_disentangle.jpg} &
\includegraphics[trim=0 384 1152 0, clip, width=0.11\textwidth]{figures/lfw_disentangle.jpg} &
\includegraphics[trim=256 384 896 0, clip, width=0.11\textwidth]{figures/lfw_disentangle.jpg} &
\includegraphics[trim=0 384 0 0, clip, width=0.11\textwidth]{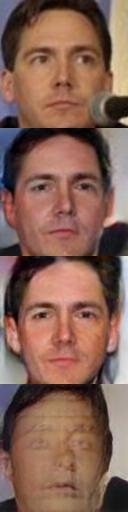} \\[-2mm]
(b) &
\includegraphics[trim=1152 256 0 128, clip, width=0.11\textwidth]{figures/lfw_disentangle.jpg} &
\includegraphics[trim=896 256 256 128, clip, width=0.11\textwidth]{figures/lfw_disentangle.jpg} &
\includegraphics[trim=128 256 1024 128, clip, width=0.11\textwidth]{figures/lfw_disentangle.jpg} &
\includegraphics[trim=0 256 1152 128, clip, width=0.11\textwidth]{figures/lfw_disentangle.jpg} &
\includegraphics[trim=256 256 896 128, clip, width=0.11\textwidth]{figures/lfw_disentangle.jpg} &
\includegraphics[trim=0 256 0 128, clip, width=0.11\textwidth]{figures/lfw_occlusions_1.jpg} \\ [-2mm]
(c) &

\includegraphics[trim=1152 128 0 256, clip, width=0.11\textwidth]{figures/lfw_disentangle.jpg} &
\includegraphics[trim=896 128 256 256, clip, width=0.11\textwidth]{figures/lfw_disentangle.jpg} &
\includegraphics[trim=128 128 1024 256, clip, width=0.11\textwidth]{figures/lfw_disentangle.jpg} &
\includegraphics[trim=0 128 1152 256, clip, width=0.11\textwidth]{figures/lfw_disentangle.jpg} &
\includegraphics[trim=256 128 896 256, clip, width=0.11\textwidth]{figures/lfw_disentangle.jpg} &
\includegraphics[trim=0 128 0 256, clip, width=0.11\textwidth]{figures/lfw_occlusions_1.jpg} \\[-2mm]
(d) &
\includegraphics[trim=1152 0 0 384, clip, width=0.11\textwidth]{figures/lfw_disentangle.jpg} &
\includegraphics[trim=896 0 256 384, clip, width=0.11\textwidth]{figures/lfw_disentangle.jpg} &
\includegraphics[trim=128 0 1024 384, clip, width=0.11\textwidth]{figures/lfw_disentangle.jpg} &
\includegraphics[trim=0 0 1152 384, clip, width=0.11\textwidth]{figures/lfw_disentangle.jpg} &
\includegraphics[trim=256 0 896 384, clip, width=0.11\textwidth]{figures/lfw_disentangle.jpg} &
\includegraphics[trim=0 0 0 384, clip, width=0.11\textwidth]{figures/lfw_occlusions_1.jpg} \\[-1mm]
& (1) & (2) & (3) & (4) & (5) & (6) \\[-4mm]
\end{tabular}
\end{center}
\figvspace \vspace{1mm}
\caption{\textbf{Feature disentanglement visualization on LFW}. We show input face ${\bf{x}}_{h}$ (a), reconstructed face from identity features $Dec({\bf{f}}_h, {\bf{0}})$ (c), non-identity features $Dec({\bf{0}}, {\bf{z}}_h)$ (d), and both (b). Feature disentanglement effectively normalizes faces in (c), and preserves pose($1$,$2$), illumination($3$), expression($4$), and occlusions($5$,$6$) in (d).} 
\label{fig:lfw}
\end{figure}

\SubSection{Ablation Study}
\Paragraph{Effects of Disentangled Feature Learning}
First, we evaluate the effects of disentangled feature learning by visualizing the disentangled identity features ${\bf{f}}_h$ and non-identity features ${\bf{z}}_h$ through our trained $Dec$. 
As shown in Fig.~\ref{fig:lfw}, the fusing of ${\bf{f}}_h$ and ${\bf{z}}_h$ can successfully recover the original image. 
The identity features alone can generate an identity-preserved frontal face, while the PIE variations and background information are captured in the non-identity features.
This suggests our framework effectively disentangles identity and non-identity features.

\begin{figure}[t]
\begin{center}
\begin{tabular}{@{}c@{\hspace{0.5mm}}c@{\hspace{1mm}}c@{\hspace{0.5mm}}c@{\hspace{1mm}}c@{\hspace{0.5mm}}c@{}}
\includegraphics[trim=1152 640 0 0, clip, width=0.11\textwidth]{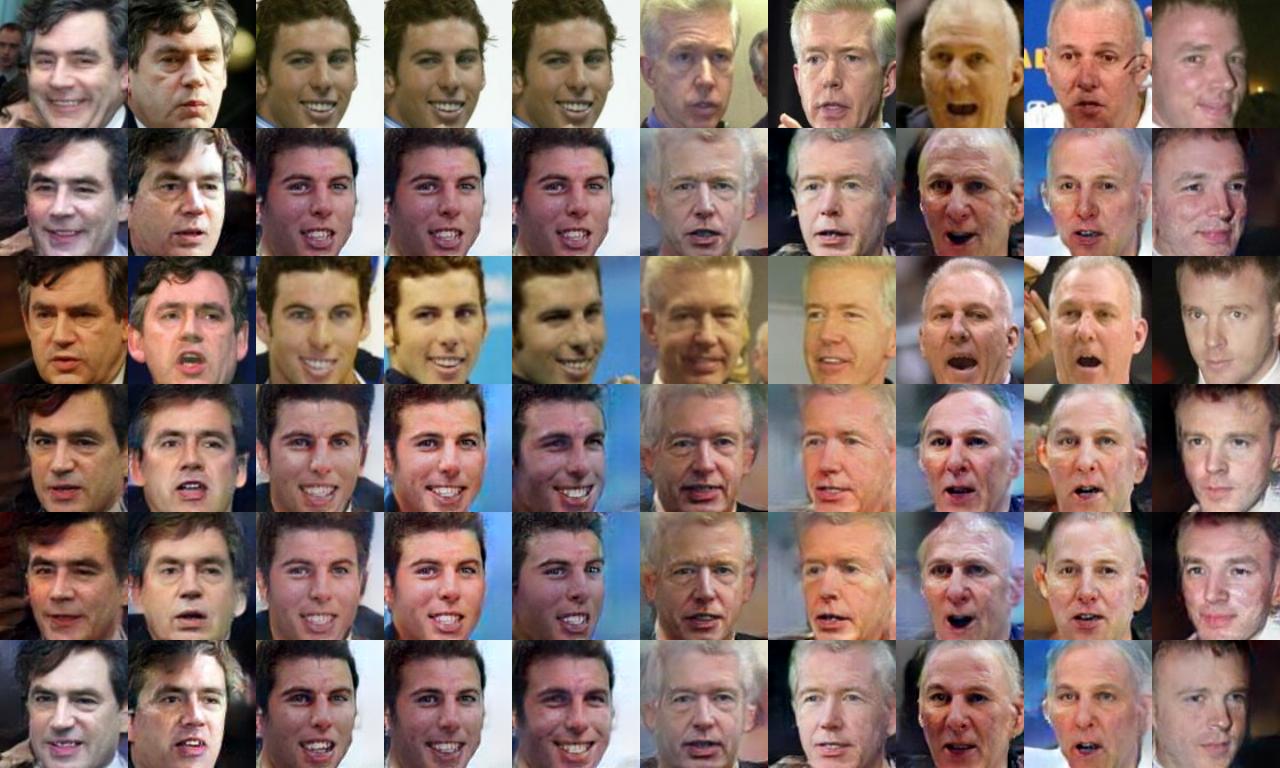} & 
\includegraphics[trim=1152 384 0 256, clip, width=0.11\textwidth]{figures/transfer/temp_9.jpg} &
\includegraphics[trim=1024 640 128 0, clip, width=0.11\textwidth]{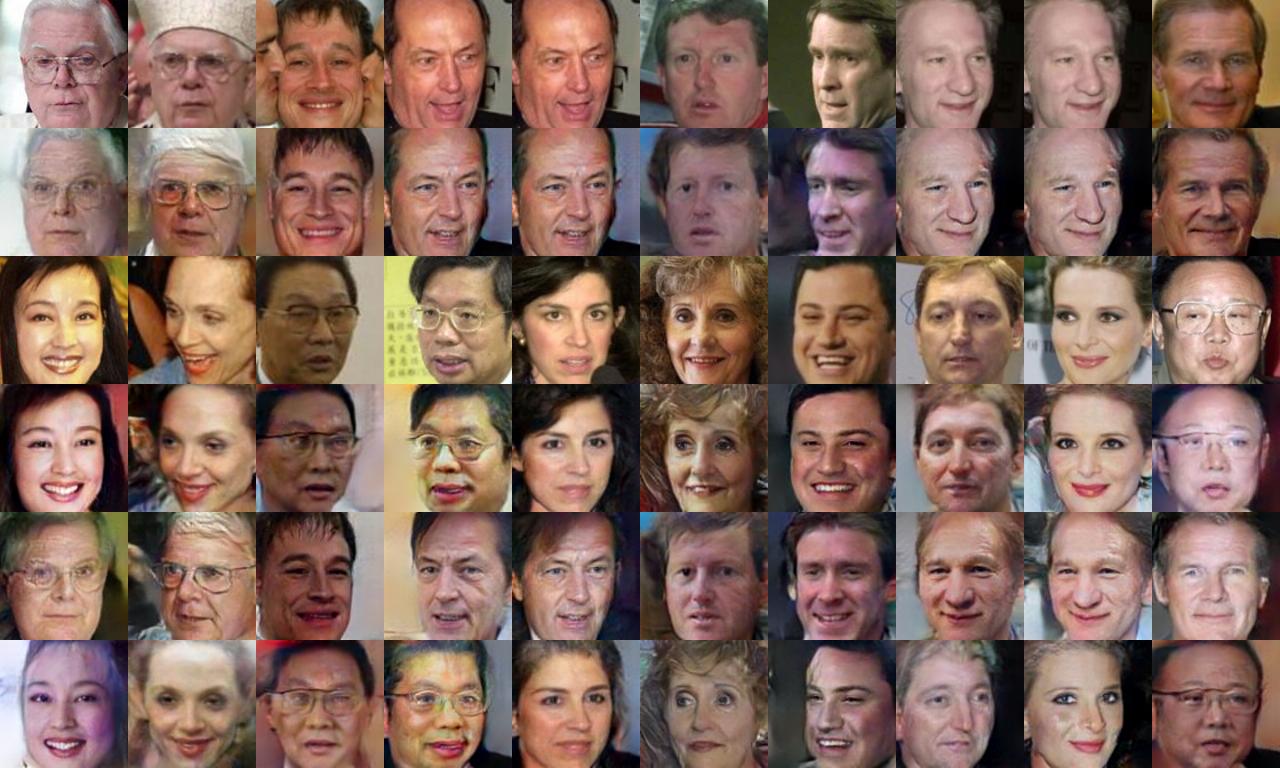} & 
\includegraphics[trim=1024 384 128 256, clip, width=0.11\textwidth]{figures/transfer/temp_36.jpg} &
\includegraphics[trim=128 640 1024 0, clip, width=0.11\textwidth]{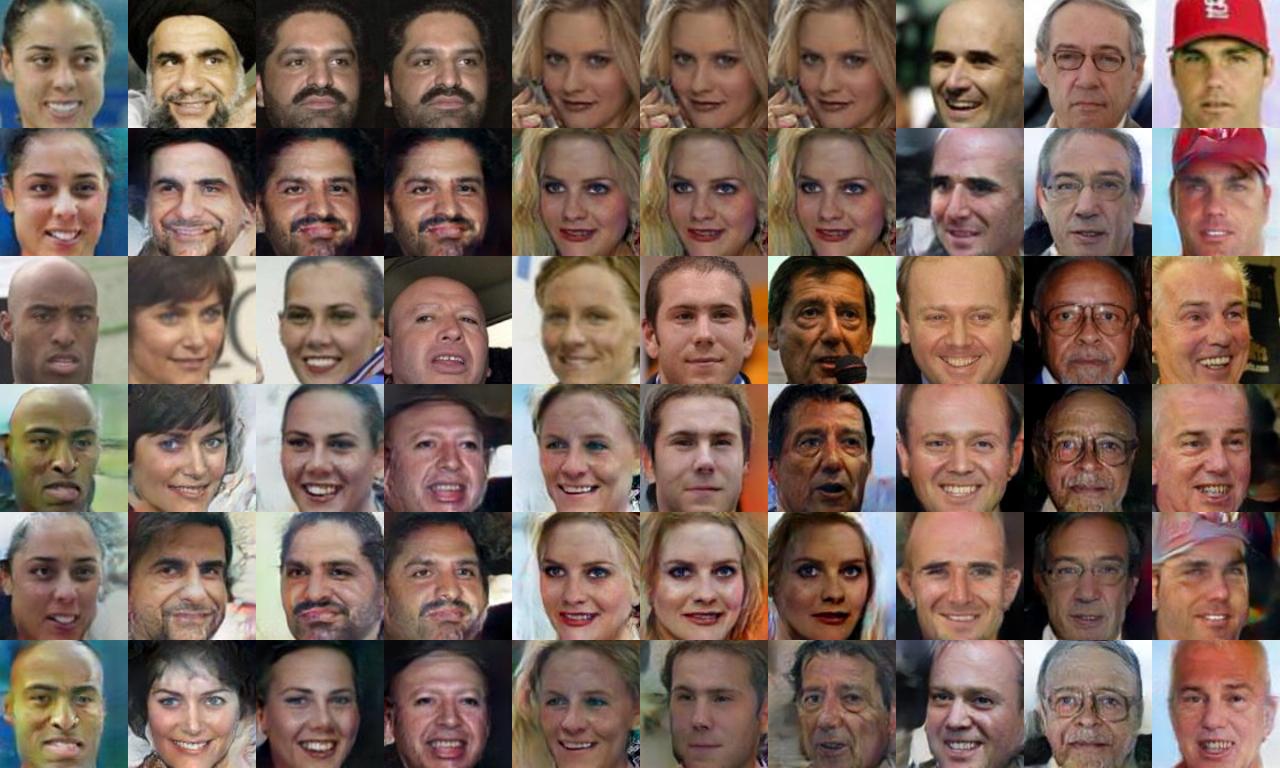} & 
\includegraphics[trim=128 384 1024 256, clip, width=0.11\textwidth]{figures/transfer/temp_92.jpg} \\[-1mm]
\includegraphics[trim=1152 128 0 512, clip, width=0.11\textwidth]{figures/transfer/temp_9.jpg} & 
\includegraphics[trim=1152 0 0 640, clip, width=0.11\textwidth]{figures/transfer/temp_9.jpg} &
\includegraphics[trim=1024 128 128 512, clip, width=0.11\textwidth]{figures/transfer/temp_36.jpg} & 
\includegraphics[trim=1024 0 128 640, clip, width=0.11\textwidth]{figures/transfer/temp_36.jpg} &
\includegraphics[trim=128 128 1024 512, clip, width=0.11\textwidth]{figures/transfer/temp_92.jpg} & 
\includegraphics[trim=128 0 1024 640, clip, width=0.11\textwidth]{figures/transfer/temp_92.jpg} \\ [-1mm]
\multicolumn{2}{c}{(a)} & \multicolumn{2}{c}{(b)} & \multicolumn{2}{c}{(c)}  \\ [-2mm]
\end{tabular}
\end{center}
\figvspace
\caption{\textbf{Feature transfer visualization between two images of same subject (a) or different subjects (b, c)}. In each set of examples, the top row shows the original images and the bottom shows the transferred images. The transferred image keeps the source's (the above image) identity and the target's (the diagonal image) attributes.}
\label{fig:transfer}
\end{figure}

Our feature disentanglement framework can also be applied for feature transfer. Given two images from either the same or different subjects, our model generates identity features as ${\bf{f}}_{h_1}$, ${\bf{f}}_{h_2}$ and non-identity features as ${\bf{z}}_{h_1}$, ${\bf{z}}_{h_2}$. We perform feature transfer from one image to the other as: $Dec({\bf{f}}_{h_1}, {\bf{z}}_{h_2})$ and $Dec({\bf{f}}_{h_2}, {\bf{z}}_{h_1})$. Fig.~\ref{fig:transfer} shows some examples where our feature transfer can keep the original image's identity and change the attributes (PIE) accordingly. 

\begin{table}[t]
\begin{center}
\small
\caption{\textbf{Face verification on super-resolved / normalized faces from LFW}. 
The $Enc\_H$ achieves $99.5\%$ on faces with the original resolutions.}
\label{table:lfw_ver}
\resizebox{0.8\linewidth}{!}{
\begin{tabular}{@{}l@{\hspace{1mm}}|c|c@{\hspace{1mm}}c@{}}
\hline
\multirow{2}{*}{Method} & $8\times$  & RSA  \\ 
                        & Acc./PSNR  & Acc./PSNR \\\hline
$Enc\_H$ & $86.6\%$/-  & $70.8\%$/- \\ 
VDSR (CVPR'$16$) + $Enc\_H$  & $85.7\%$/$26.63$  & $69.4\%$/$25.49$ \\
SRResNet (CVPR'$17$) + $Enc\_H$ & $86.2\%$/$27.46$  & $68.6\%$/$24.85$ \\
FSRNet (CVPR'$18$) + $Enc\_H$ & $89.7\%$/$28.27$ & $69.4\%$/$25.25$ \\
FSRGAN (CVPR'$18$) + $Enc\_H$ & $86.7\%$/$26.36$ & $67.0\%$/$23.94$ \\ \hline
FAN ({\emph{i.e.}}, normalized image) + $Enc\_H$ & $91.9\%$/-  & $76.8\%$/- \\
FAN ({\emph{i.e.}}, $Enc\_L$) & $\textbf{95.2}\%$/- & $\textbf{82.4}\%$/- \\ \hline
\end{tabular}}\figvspace
\end{center}
\end{table}

\Paragraph{Effects of Joint Learning}
We conduct verification tests to demonstrate the effects of jointly learning both face recognition and normalization, compared with $Enc\_H$ that is only trained for face recognition, and state-of-the-art SR methods such as VDSR~\cite{VDSR_CVPR16}, SRResNet~\cite{SRGAN_CVPR17} and FSRNet/FSRGAN~\cite{CT-FSRNet-2018}, that are only trained for face hallucination.
With the standard verification protocol of LFW, we down-sample the $6,000$ test pairs to $16\times16$ with a $8\times$ scale factor, and upscale back to $128\times128$ via bicubic interpolation. 
Unlike~\cite{zhang2018facesr} that retrains recognition network on the super-resolved images, we directly use $Enc\_H$ to evaluate the accuracy ({\emph{i.e.}}, Acc.) of different face SR methods.

Since FAN handles face hallucination and frontalization simultaneously, it is not suitable to use pixel-wise evaluation metrics ({\emph{e.g.}}, PSNR) to evaluate our method. 
Instead, we compare face verification performance. 
From the results in Tab.~\ref{table:lfw_ver}, we have three observations.  
1) The recognition performance drops significantly when processing LR faces. 
2) After incorporating SR methods, FSRNet~\cite{CT-FSRNet-2018} achieves better verification ({\emph{i.e.}}, Acc.) and SR ({\emph{i.e.}}, PSNR) results than other SR methods. Our FAN achieves the best results among all SOTA SR methods. 
3) It is more effective to learn identity features from LR input (Our $Enc\_L$) than performing face hallucination and recognition on the normalized faces (Our normalized image + $Enc\_H$).

\Paragraph{Effects of Random Scale Augmentation}
We further conduct tests to demonstrate the effects of our RSA strategy under the same experimental setting as the above study.
The only difference is that we randomly down-sample the $6,000$ test pairs to the resolution interval between $8$$\times$$8$ and $32$$\times$$32$, \emph{i.e.}, the scale factors from $16$$\times$ to $4$$\times$.
The extreme LR $8$$\times$$8$ and various/unknown resolutions existed in test images are more common than a larger and fixed resolution case.
As shown in Tab.~\ref{table:lfw_ver}, we have three observations.
1) Since images with much lower resolutions ({\emph{e.g.}}, $8$$\times$$8$) are evaluated, the baseline $Enc\_H$ is much lower than the $8\times$ case.
2) Since all of the SR methods are trained with specific scales ({\emph{e.g.}}, $8\times$), they are not suitable for test images with varying resolutions.
3) Our FAN is much better than other methods in this case, which demonstrates the effects of RSA and our FAN is more practical in real-world scenarios.

\begin{figure} [t!]
\begin{center}
\includegraphics[width=0.75\linewidth]{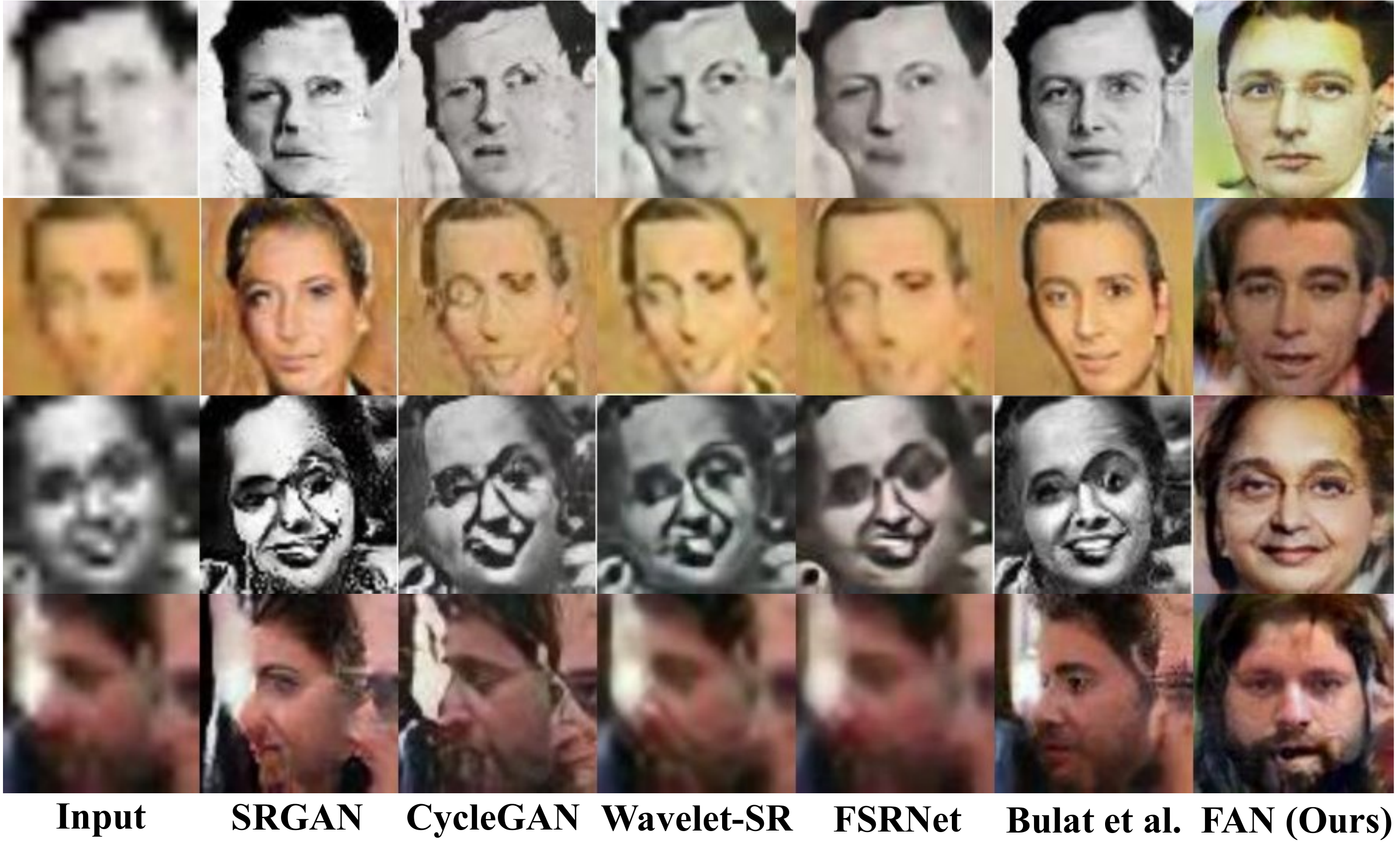} \vspace{-6mm}
\end{center}
\caption{\textbf{Face SR/normalization on WIDER FACE}.}
\figvspace
\label{fig:widerface_comp1}
\end{figure}

\begin{figure} [t!]
\begin{center}
\includegraphics[width=0.75\linewidth]{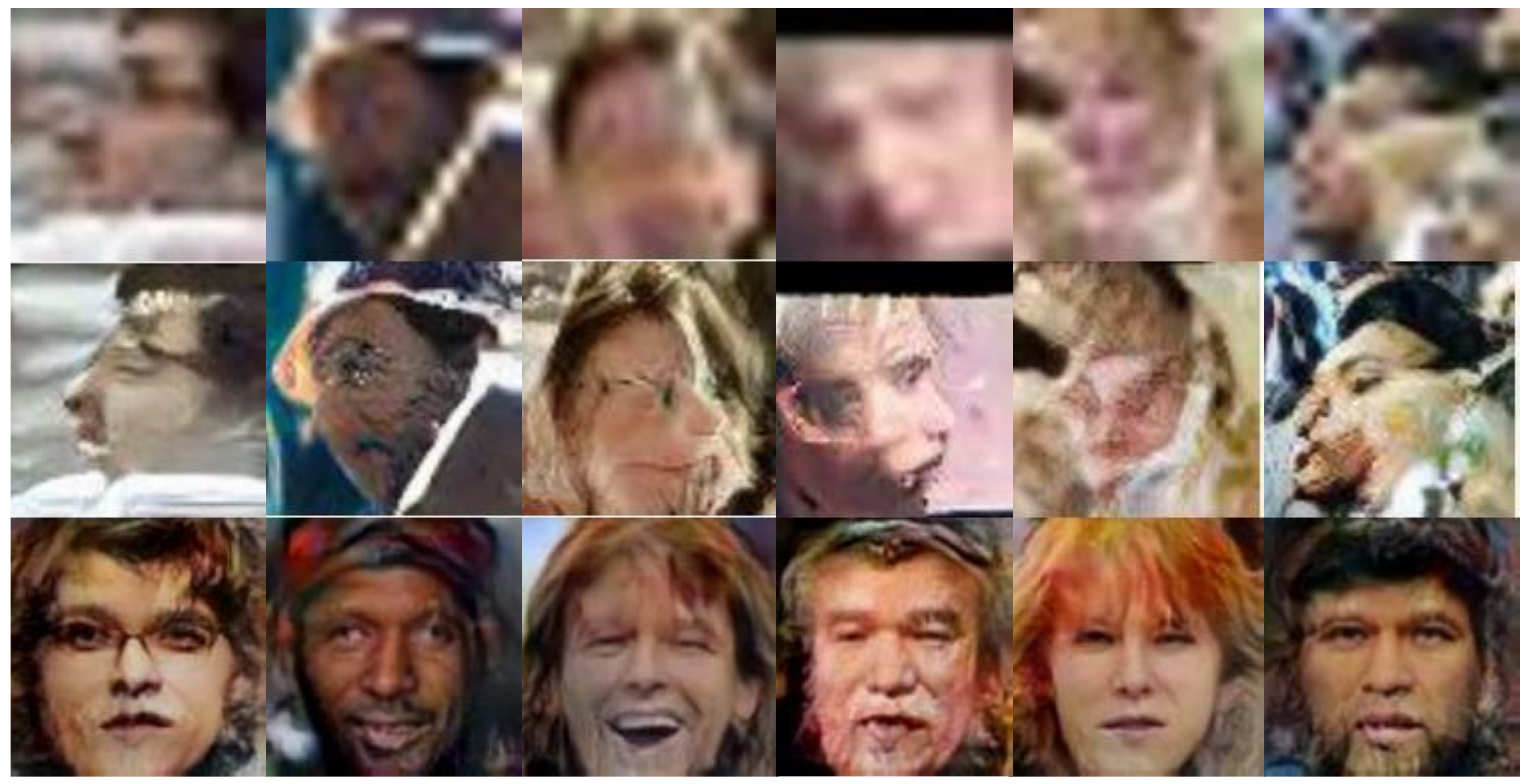} \vspace{-6mm}
\end{center}
\caption{\textbf{Face SR/normalization on heavily distorted images from WIDER FACE}. Top: input images. Middle: Bulat \emph{et al.}~\cite{bulatyang2018learn}. Bottom: our FAN ({\emph{i.e.}}, $Dec({\bf{f}}_l, {\bf{0}})$).}
\figvspace
\label{fig:widerface_comp2}
\end{figure}

\SubSection{Comparisons with SOTA Methods}
In this section, we conduct extensive comparisons with SOTA methods both quantitatively and qualitatively. 
First, we show the hallucination ability of our method on recovering HR faces from heavily distorted faces on WIDER FACE~\cite{yang2016widerface}.
Second, we demonstrate the ability for preserving the identity information via verification/identification comparisons on QMUL-SurFace~\cite{cheng2018qmulface} and SCface~\cite{Grgic2011scface}.

\Paragraph{Comparisons on WIDER FACE}
First, we qualitatively compare our FAN with CycleGAN~\cite{CycleGAN2017}, SRGAN~\cite{SRGAN_CVPR17}, Wavelet-SRNet~\cite{huang2017waveletsr}, FSRNet~\cite{CT-FSRNet-2018} and Bulat \emph{et al.}~\cite{bulatyang2018learn} on WIDER FACE.
Figure~\ref{fig:widerface_comp1} illustrates the recovered images, where the results of the competitors are all imported directly from~\cite{bulatyang2018learn}. 
As we can see, both our FAN and~\cite{bulatyang2018learn} perform well on recovering valid faces in these cases.
However, there are two differences between ours and the competitors.
First, FAN is trained on refined-MS1M for joint face recognition and normalization, and we do \textbf{NOT} finetune our model with the WIDE FACE data, while the competitors are directly trained with $\sim50K$ images from WIDER FACE.
Second, FAN generates $128\times128$ HR images, while the competitors super-resolve to $64\times64$, a relatively \textbf{easier} task.

We further test our method on some heavily distorted images that~\cite{bulatyang2018learn} fails to recover meaningful HR faces, and show the results in Fig.~\ref{fig:widerface_comp2}.
Thank to our powerful encoder and decoder networks, even when dealing with extremely low-quality faces, our method still recovers valid faces that are much clear than~\cite{bulatyang2018learn}.
It is an open question that whether we shall recover a face from a heavily distorted image. 
To address this issue, one could estimate the face quality and determine when normalization should be applied, which is not in the scope of our work or most prior face SR methods but a good direction for future work.

\begin{figure}[t]
\begin{center}
\begin{tabular}{@{}c@{}c@{}c@{}c@{}c@{}c@{}c@{}c@{}cc@{}c@{}c@{}c@{}c@{}c@{}}
(a) & 
\includegraphics[trim=0 128 640 0, clip, width=0.11\textwidth]{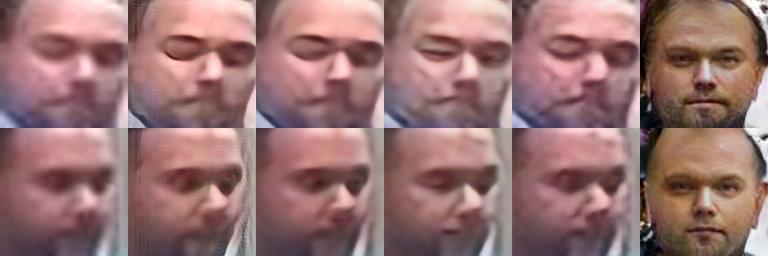} &
\includegraphics[trim=128 128 512 0, clip, width=0.11\textwidth]{figures/013_2276_cam2_1.jpg} &
\includegraphics[trim=256 128 384 0, clip, width=0.11\textwidth]{figures/013_2276_cam2_1.jpg} &
\includegraphics[trim=384 128 256 0, clip, width=0.11\textwidth]{figures/013_2276_cam2_1.jpg} &
\includegraphics[trim=512 128 128 0, clip, width=0.11\textwidth]{figures/013_2276_cam2_1.jpg} &
\includegraphics[trim=640 128 0 0, clip, width=0.11\textwidth]{figures/013_2276_cam2_1.jpg} \\ [-2mm]
&
\includegraphics[trim=0 0 640 128, clip, width=0.11\textwidth]{figures/013_2276_cam2_1.jpg} &
\includegraphics[trim=128 0 512 128, clip, width=0.11\textwidth]{figures/013_2276_cam2_1.jpg} &
\includegraphics[trim=256 0 384 128, clip, width=0.11\textwidth]{figures/013_2276_cam2_1.jpg} &
\includegraphics[trim=384 0 256 128, clip, width=0.11\textwidth]{figures/013_2276_cam2_1.jpg} &
\includegraphics[trim=512 0 128 128, clip, width=0.11\textwidth]{figures/013_2276_cam2_1.jpg} &
\includegraphics[trim=640 0 0 128, clip, width=0.11\textwidth]{figures/013_2276_cam2_1.jpg} \\ [-1mm]
(b) &
\includegraphics[trim=0 128 640 0, clip, width=0.11\textwidth]{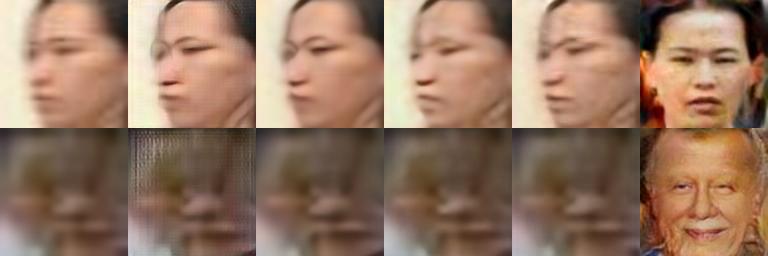} &
\includegraphics[trim=128 128 512 0, clip, width=0.11\textwidth]{figures/017_2047_cam1_1.jpg} &
\includegraphics[trim=256 128 384 0, clip, width=0.11\textwidth]{figures/017_2047_cam1_1.jpg} &
\includegraphics[trim=384 128 256 0, clip, width=0.11\textwidth]{figures/017_2047_cam1_1.jpg} &
\includegraphics[trim=512 128 128 0, clip, width=0.11\textwidth]{figures/017_2047_cam1_1.jpg} &
\includegraphics[trim=640 128 0 0, clip, width=0.11\textwidth]{figures/017_2047_cam1_1.jpg} \\ [-2mm]
&
\includegraphics[trim=0 0 640 128, clip, width=0.11\textwidth]{figures/017_2047_cam1_1.jpg} &
\includegraphics[trim=128 0 512 128, clip, width=0.11\textwidth]{figures/017_2047_cam1_1.jpg} &
\includegraphics[trim=256 0 384 128, clip, width=0.11\textwidth]{figures/017_2047_cam1_1.jpg} &
\includegraphics[trim=384 0 256 128, clip, width=0.11\textwidth]{figures/017_2047_cam1_1.jpg} &
\includegraphics[trim=512 0 128 128, clip, width=0.11\textwidth]{figures/017_2047_cam1_1.jpg} &
\includegraphics[trim=640 0 0 128, clip, width=0.11\textwidth]{figures/017_2047_cam1_1.jpg} \\ [-1mm]
 & \small{Input} & \small{FSRGAN} & \small{FSRNet} & \small{SRResNet} & \small{VDSR} & \small{FAN (Ours)}\\ [-2mm]
\end{tabular}
\end{center}
\figvspace
\caption{\textbf{Face SR/normalization on the verification set of QMUL-SurvFace}. (a): a face pair of the same subject. (b): a face pair of different subjects.}
\vspace{-3mm}
\label{fig:qmul}
\end{figure}

\Paragraph{Comparisons on QMUL-Surv}
QMUL-Surv includes very LR faces captured with surveillance cameras. 
It is a very challenging dataset as most of the faces are hardly visible.
We compare our framework with SOTA face SR methods~\cite{VDSR_CVPR16, SRGAN_CVPR17, CT-FSRNet-2018} to evaluate the performance on real world surveillance data. 
As shown in Fig.~\ref{fig:qmul}, previous works struggle to recover the high-frequency information from the input LR faces. 
In contrast, our FAN can consistently generate a high-quality frontal face that recovers identity information. 
In addition to this qualitative evaluation, we also conduct face verification evaluation on the super-resolved/normalized face images. 
As shown in Tab.~\ref{tab:qmul}, our FAN performs better than previous SR methods. 
We have also evaluated the performance of using $Enc\_L$ to extract features directly from the LR inputs. 
The results in Tab.~\ref{tab:qmul} indicates that it is more effective to learn features rather than performing super-resolution for LR face recognition, which is consistent with the observation in Tab.~\ref{table:lfw_ver}.

\begin{table}[t]
\begin{center}
\small
\caption{\textbf{Face verification results on QMUL-Surv super-resolved faces evaluated with $Enc\_H$ except the last row.}}
\label{tab:qmul}
\begin{tabular}{@{}l@{\hspace{0.6mm}}|c@{\hspace{0.6mm}}c@{\hspace{0.6mm}}c@{\hspace{0.6mm}}c@{\hspace{0.6mm}}|c@{\hspace{0.6mm}}|c@{}}
\hline
\multirow{2}{*}{Method} & \multicolumn{4}{c|}{TAR($\%$)@FAR} & \multirow{2}{*}{AUC} & Mean \\
 &  $30\%$ & $10\%$ & $1\%$ & $0.1\%$ & & Acc ($\%$) \\ \hline
VDSR~\cite{VDSR_CVPR16} & $61.03$ & $35.32$ & $8.89$ & $3.10$ & $71.02$ & $65.64$ \\
SRResNet~\cite{SRGAN_CVPR17} & $61.81$ & $34.03$ & $8.36$ & $2.07$ & $71.00$ & $65.94$\\
FSRNet~\cite{CT-FSRNet-2018} & $59.92$ & $33.10$ & $7.84$ & $1.93$ & $70.09$ & $64.96$ \\
FSRGAN~\cite{CT-FSRNet-2018} & $56.03$ & $30.91$ & $8.45$ & $2.66$ & $67.93$ & $63.06$ \\ \hline
FAN (norm. face) & $62.31$ & $36.64$ & $11.89$ & $\bf 3.70$ & $71.66$ & $66.32$ \\
FAN ($Enc\_L$) & $\bf 71.30$ & $\bf 44.59$ & $\bf 12.94$ & $2.75$ & $\bf 76.94$ & $\bf 70.88$ \\ \hline
\end{tabular}
\end{center}
\figvspace
\end{table}

\Paragraph{Comparisons on SCface}
SCface defines face identification with unpaired HR and LR faces. 
It mimics the real-world surveillance watch-list problem, where the gallery contains HR faces and the probe consists of LR faces captured from surveillance cameras.
The HR and LR images do not have correspondence in the pixel domain, which is difficult for previous face SR methods that requires pixel-level correspondence. 
It is not a problem for FAN as we regularize the model training in the disentangled feature space. 
Following~\cite{lu2018sr}, we conduct experiments on the daytime data only. 
The first $80$ subjects are used for testing and the rest $50$ subjects are for fine-tuning the second stage of our method. 
In addition to the unpaired HR and LR images in the training set, we also perform RSA to generate LR images from the HR images for model fine-tuning. 

\begin{table}[t!]
\begin{center}
\small
\caption{\textbf{Rank-$1$ performance of face identification on SCface testing set}. `-FT' means fine-tuning with SCface training set. Most compared results are cited from~\cite{lu2018sr} except ArcFace that we pretrained on refined MS$1$M.}
\label{tab:SCface}
\begin{tabular}{l|ccc|c}
\hline
Distance $\to$ & d$1$ & d$2$ & d$3$ & avg. \\ \hline
LDMDS~\cite{yang2018discriminative} & $62.7$ & $70.7$ & $65.5$ & $66.3$ \\
LightCNN~\cite{wu2018light} & $35.8$ & $79.0$ & $93.8$ & $69.5$ \\
Center Loss~\cite{wen2016discriminative} & $36.3$ & $81.8$ & $94.3$ & $70.8$ \\
ArcFace (ResNet$50$)~\cite{deng2019arcface} & $48.0$ & $92.0$ & $\bf{99.3}$ & $79.8$ \\  \hline
LightCNN-FT & $49.0$ & $83.8$ & $93.5$ & $75.4$ \\
Center Loss-FT & $54.8$ & $86.3$ & $95.8$ & $79.0$ \\
ArcFace (ResNet$50$)-FT & $67.3$ & $93.5$ & $98.0$ & $86.3$ \\
DCR-FT~\cite{lu2018sr} & $73.3$ & $93.5$ & $98.0$ & $88.3$ \\ \hline
FAN & $62.0$ & $90.0$ & $94.8$ & $82.3$    \\
FAN-FT (no RSA)& $68.5$ & $92.3$ & $97.8$ & $86.2$  \\ 
FAN-FT (no $Dec$) & $73.0$ & $94.0$ & $97.8$ & $88.3$ \\ 
FAN-FT & $\bf{77.5}$ & $\bf{95.0}$ & ${98.3}$ & $\bf{90.3}$  \\ 
\hline 
\end{tabular}
\figvspace
\end{center}
\end{table}

We mainly compare with DCR~\cite{lu2018sr} as it achieved SOTA results on SCface.
As far as we know, almost all SOTA face recognition methods have not evaluated on SCface.
For fair comparison, we implemented ArcFace~\cite{deng2019arcface} using the same backbone and also finetuned on SCface training set.
As shown in Tab.~\ref{tab:SCface}, 
our FAN achieves the best results among all other methods that are not finetuned on SCface.
After finetuning on SCface with RSA, we achieve new SOTA results.
Note that DCR proposed to use decoupled training that learns feature mapping for faces at {\it each} resolution, and the resolution information is assumed to be given in the testing stage. However, such resolution information is often unavailable in practice. 
Nevertheless, our method still outperforms DCR by a large margin even though we do {\bf NOT} use the resolution information at the testing stage. 
From the gap between FAN-FT and FAN-FT (no RSA), we can see the effectiveness of RSA for surveillance FR. 
We also conducted ablative experiments by removing the $Dec$. 
No $Dec$ means that we only perform feature-level similarity regularization in the second stage. 
The results of FAN-FT (no $Dec$) suggests that joint learning of face normalization can help feature learning for FR.

\begin{figure}[t]
\begin{center}
\includegraphics[trim=50 60 260 95, clip,width=0.75\textwidth]{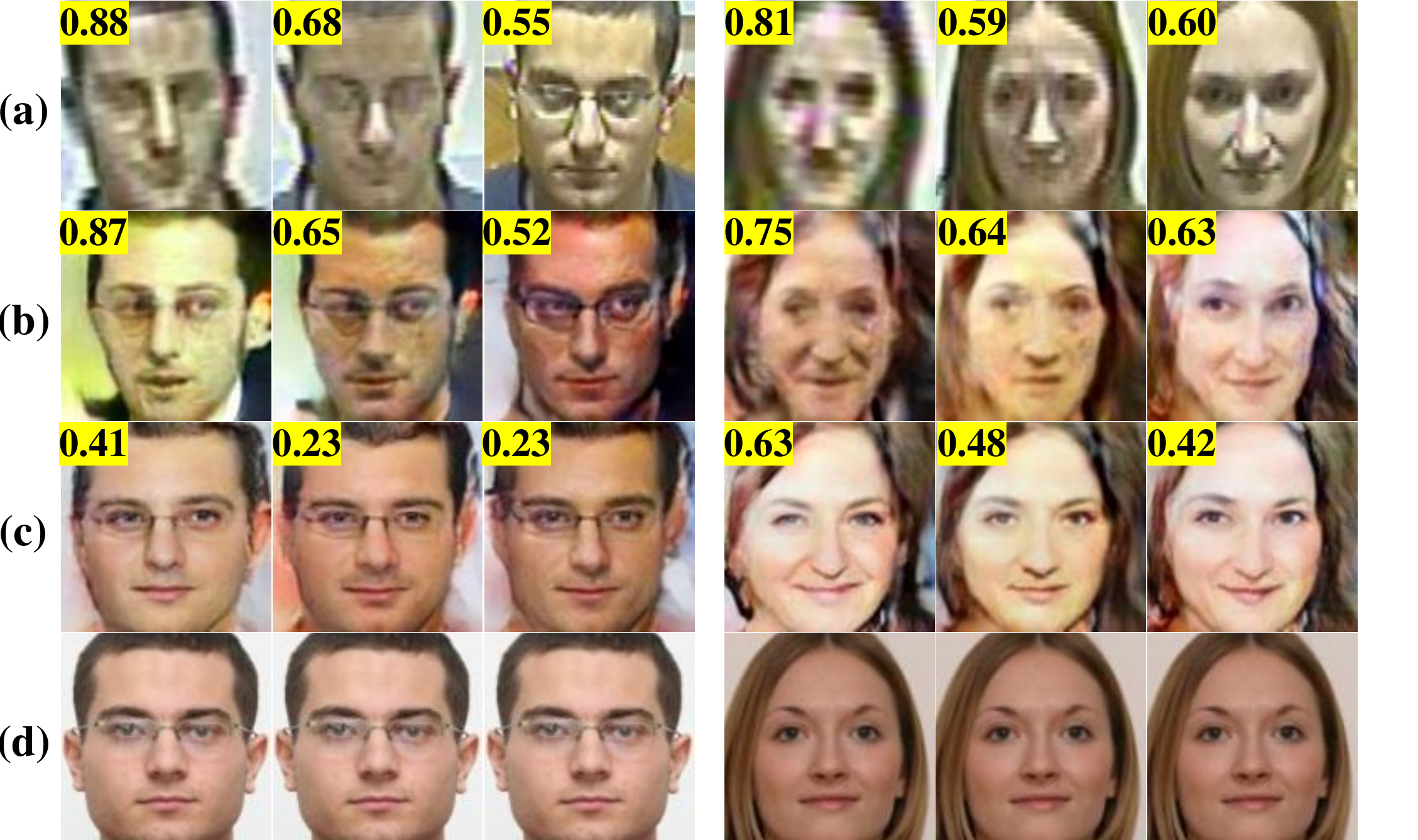}
\end{center}
\figvspace
\caption{\textbf{Face normalization on SCface testing set}. (a) input images at three resolutions. (b) normalized faces generated by $Enc\_H$ and $Dec$. (c) normalized faces generated by $Enc\_L$ and $Dec$. (d) HR gallery images. The number indicates the feature distance between the input / normalized face and the gallery HR face.}
\label{fig:scface}
\end{figure}

Figure~\ref{fig:scface} shows our face normalization results on SCface testing set.
Our method can generate high-quality face images that recover the identity information from the input faces with various resolutions. 
The comparison between the results generated by $Enc\_H$ and $Enc\_L$ validates the effectiveness of feature adaptation in our second stage both quantitatively and qualitatively.

\Paragraph{Time Complexity}
Our framework adopts an encoder-decoder ({\emph{i.e.}}, $Enc\_L$ and $Dec$) structure, it takes $\sim$$0.011$s to extract \textit{compact} identity features, and another $\sim$$0.005$s to recover a $128$$\times$$128$ HR image on Titan X GPU, which is comparable to $0.012$s for FSRNet~\cite{CT-FSRNet-2018} on the same hardware, and much faster than $3.84$s for CBN~\cite{CBN_ECCV16}, $8$ min.~for~\cite{facesr1_ijcv07} and $15$ min.~for~\cite{facesr2_cvpr15}.
In general, compared with the SOTA methods, our FAN is a better choice in surveillance scenarios considering both visualization and recognition.

\Section{Conclusions}\label{section:5}
This paper proposes a Feature Adaptation Network (FAN) for surveillance face recognition and normalization.
Despite the great improvement in face recognition and super-resolution, the applications in surveillance scenario is less studied. 
We aim to bridge this gap. 
FAN consists of two stages: feature disentanglement learning and feature adaptation.
By first disentangling the face features into identity and non-identity components, it enables our adaptation network to impose both feature-level and image-level similarity regularizations.
Such framework is suitable for both paired and unpaired training, which overcomes the limit by previous face SR methods that require paired training data.
The proposed Random Scale Augmentation (RSA) is very effective in handling the various resolutions in surveillance imagery. 
We achieved SOTA face recognition and normalization results even from very low quality inputs.

\bibliographystyle{splncs}
\bibliography{egbib}

\end{document}